\documentclass[conference]{IEEEtran}
\IEEEoverridecommandlockouts
\usepackage{cite}
\usepackage{amsmath,amssymb,amsfonts}
\usepackage{algorithmic}
\usepackage{graphicx}
\usepackage{textcomp}
\usepackage{xcolor}
\def\BibTeX{{\rm B\kern-.05em{\sc i\kern-.025em b}\kern-.08em
    T\kern-.1667em\lower.7ex\hbox{E}\kern-.125emX}}

\usepackage{amsmath}
\usepackage{booktabs}

\begin{document}

\graphicspath{{./figure}}
\DeclareGraphicsExtensions{.pdf,.jpeg,.png,.jpg,.bmp}

\title{Follow-Your-MultiPose: Tuning-Free Multi-Character Text-to-Video Generation via Pose Guidance}

% \author{\IEEEauthorblockN{Author(s) Name(s)}
% \IEEEauthorblockA{Author Affiliation(s)}}

\author{
\IEEEauthorblockN{Beiyuan Zhang$^{1\dag}$ , Yue Ma$^{2\dag}$, Chunlei Fu$^{1}$, Xinyang Song$^{3,4}$, Zhenan Sun$^{3,4}$, Ziqiang Li$^{5}$}
\IEEEauthorblockA{$^1$ School of Big Data and Software Engineering, Chongqing University, Chongqing, China}
\IEEEauthorblockA{$^2$ Computer Science Engineer, The Hong Kong University of Science and Technology, Hong Kong}
\IEEEauthorblockA{$^3$ School of Artificial Intelligence, University of Chinese Academy of Sciences, Beijing, China}
\IEEEauthorblockA{$^4$ New Laboratory of Pattern Recognition (NLPR), Institute of Automation, Chinese Academy of Sciences, Beijing, China}
\IEEEauthorblockA{$^5$ Department of Computer Science and Engineering, Shanghai Jiao Tong University, Shanghai, China}\vspace{-0.25em}
\thanks{${\dag}$ These authors contributed equally to this work.\par
${\ast}$ Corresponding author: clfu@cqu.edu.cn}
\IEEEauthorblockA{Beiyuan.Zhang@stu.cqu.edu.cn, mayuefighting@gmail.com, clfu@cqu.edu.cn,\\ xinyang.song@cripac.ia.ac.cn, znsun@nlpr.ia.ac.cn, ziqiangli@sjtu.edu.cn}
}

% \and
% \and
% \and
% \and
% \and

\maketitle

%%模块的命名，语法，结构

\begin{abstract}
Text-editable and pose-controllable character video generation is a challenging but prevailing topic with practical applications. However, existing approaches mainly focus on single-object video generation with pose guidance, ignoring the realistic situation that multi-character appear concurrently in a scenario. To tackle this, we propose a novel multi-character video generation framework in a tuning-free manner, which is based on the separated text and pose guidance. Specifically, 
we first extract character masks from the pose sequence to identify the spatial position for each generating character, and then single prompts for each character are obtained with LLMs for precise text guidance. Moreover, the spatial-aligned cross attention and multi-branch control module are proposed to generate fine-grained controllable multi-character video. The visualized results of generating video demonstrate the precise controllability of our method for multi-character generation. We also verify the generality of our method by applying it to various personalized T2I models. Moreover, the quantitative results show that our approach achieves superior performance compared with previous works.
\end{abstract}

\begin{IEEEkeywords}
Text to video generation, Pose guided video generation, Multi- character generation.
\end{IEEEkeywords}

\section{Introduction}
Controllable characters generation task takes condition signals such as text prompt, motion video, depth sequence, pose sequence or any combination of these conditions as input, and generates videos in line with them. Due to its potential application prospects in films, cartoons and short videos, controllable characters generation has attracted more and more attention in both academia and industry. Among these, character generation via text prompts and pose sequences is the most attractive approach, given its precision, validity and generalization. 

\begin{figure}[h]
  \centering
  \includegraphics[width=0.97\linewidth]{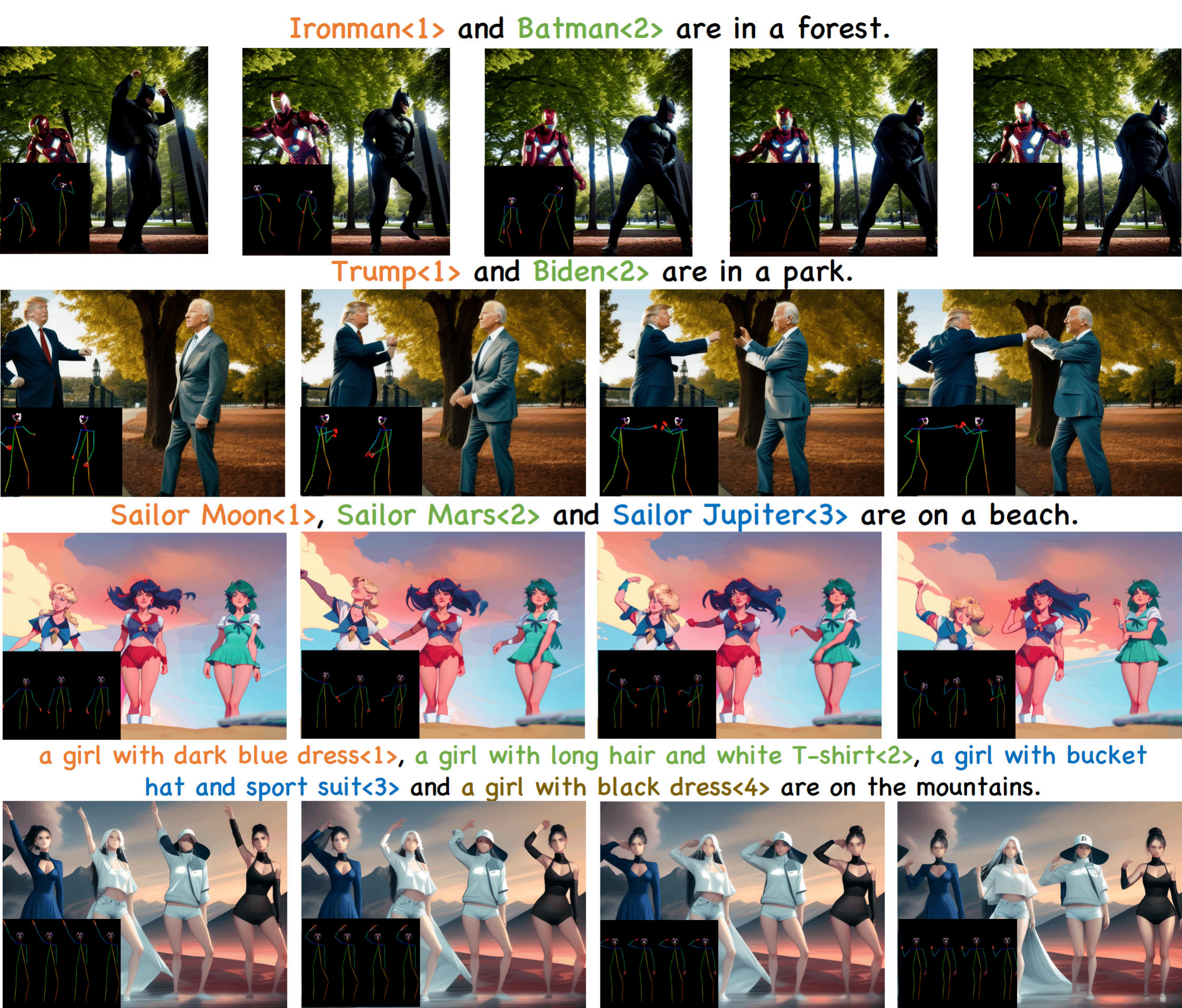}
  \caption{The visualization of multi-character generation of our method.}
  \label{pic1}
\end{figure}

In earlier studies\cite{everybody,deep,bidirectionally,exploring}, researchers have explored this area with methods based on GAN~\cite{gan}. However, the results of generation display many issues such as blurred details, semantic inconsistency, pose inconsistency and temporal instability. Recently, diffusion-based text-to-image models\cite{high,glide,hierarchical,photorealistic} have obtained tremendous improvement in the quality and stability of generation. With that, diffusion models with conditions\cite{click,stick,emoji,canvas,cove,swap,minds,dit} in the field of generation and editing spring up. Existing diffusion-based methods\cite{follow,dreampose, advancing,humansd,person} accomplish the topic of text-editable and pose-controllable character generation in both image and video generation. By their method, we can get semantic and pose well-aligned, artistic and temporal consistency generation results. However, most of the existing methods only focus on the generation of single character using prompts and they fail to generate fine-aligned multiple character according various prompts. This observation leads to a exploration:
\begin{figure*}[h]
  \vspace{-1.0em}
  \centering
  \includegraphics[width=\textwidth]{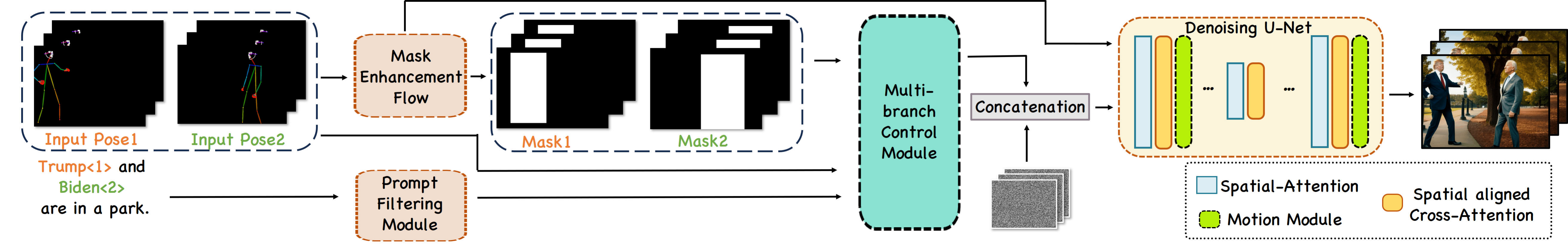}
  \caption{Framework of Follow-Your-MultiPose.}
  \label{pic2}
  \vspace{-1.5em}
\end{figure*}
\textbf{\textit{Can we achieve more fine-grained control in multi-character generation?}}

In this work, to address the limitations above, we design a tuning-free method called Follow-Your-MultiPose(FYM) to achieve multi-character text-to-video generation via pose guidance. Concretely, we first obtain character masks from the pose sequence to identify reasonable spatial positions for the multi-generating characters in one video. Then, new text prompts for every single character are 
re-extracted with large language models(i.e., LLAMA~\cite{llama}, GPT4~\cite{gpt}) for every single character from the description of multi-character, leading to precise text guidance for various characters. Moreover, spatial aligned cross-attention is proposed to integrate the different guidance of newly extracted prompts and character masks, generating more fine-grained video clips. Finally, we introduce the multi-branch control module inputted with the corresponding prompt, pose sequence and mask of each character, achieving clearer controllability for each generating character. Thanks to our tuning-free strategy, we can successfully generate multi-character videos with high temporal consistency that succeed from pre-trained models, and distinct generation styles also be achieved by loading different pre-trained checkpoints from CivitAI\cite{civitai}. The visualization of generation results is shown in Fig.~\ref{pic1}. Extensive quantitative and qualitative experiments are conducted, demonstrating the superiority of our approach compared with existing methods~\cite{follow, controlvideo, text2video, first, masactrl, tune}. 

In summary, our contributions are as follows:
\begin{itemize}
\item We propose a tuning-free method to tackle the problem of pose guided multi-character text-to-video generation. Guided by the mask mechanism, our method can generate characters at any specified spatial section in the video.

\item We propose the spatial-aligned cross-attention mechanism and multi-branch control module to achieve fine-grained controllability for multi-character video generation.

\item Extensive experiments compared with various baselines demonstrate the superiority of our approach, in terms of generation quality, text-video alignment, pose-video alignment, and temporal coherence.
\end{itemize}

\section{Method}

The overview of our FYM is shown in Fig.~\ref{pic2}, and it is composed of four parts: mask enhancement flow, prompt filtering module, spatial-aligned cross attention and multi-branch control module. We will depict them in the following sections. Notice that, our proposed method is a general tuning-free method and can be used in various text-to-video models. In this paper, we majorly follow the Animatediff\cite{animatediff} for its popularity and generalization ability.  

\subsection{Mask Enhancement Flow}
%%character location module
%%Mask Converse Module
\label{sec:Mask Converse Module.}
In mask enhancement flow, we extract and transform masks from pose sequences. The extracted masks will serve as identifiers for the designated region, allowing our method to generate characters at the specified position in one video.

The whole process is divided into three steps. As demonstrated in Fig.~\ref{pic4}, we first design a mask extractor to detect regions from pose sequences in each group. The extractor locates the first and last rows with non-zero pixels, and the first and last columns with non-zero pixels. The pixel values within this region are set to one, and others are set to zero. Each region is treated as a spatial section. The masks of the same frames number in each spatial section are divided into groups for subsequent processing. 

Next, we do mask enhancement for masks in each group. We first normalize the masks using Softmax, the sum of the pixel values at the same position of each mask is one. Then, we convert masks into three groups using interpolation method.

Group (a) is for spatial aligned cross-attention, $(h,w)$ stands for the size of the hidden states and $\left ( \frac{h}{2},\frac{w}{2} \right ), \left ( \frac{h}{4},\frac{w}{4} \right )$ are the size of intermediate dimensions in the U-Net down sampling process. Group (b) and group (c) are used for multi-branch control module, corresponding respectively to the down block samples and middle block samples of multiple ControlNet\cite{controlnet}. ControlNet\cite{controlnet} down block samples sizes are $(h,w)$, $\left ( \frac{h}{2},\frac{w}{2} \right ), \left ( \frac{h}{4},\frac{w}{4} \right )$, $\left ( \frac{h}{8},\frac{w}{8} \right )$, and the ControlNet\cite{controlnet} middle block samples size is likewise $\left ( \frac{h}{8},\frac{w}{8} \right )$.

\begin{figure*}[h]
  \vspace{-1.0em}
  \centering
  \includegraphics[width=0.95\linewidth]{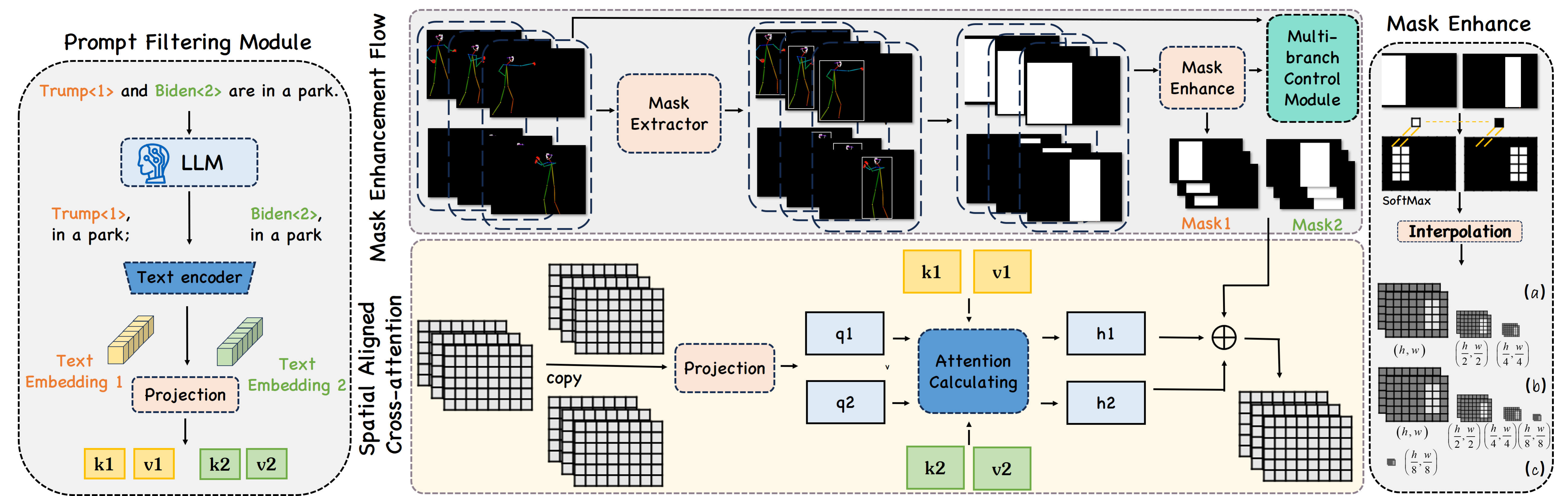}
  \caption{Details of each module, including prompt filtering module, mask enhancement flow and spatial aligned cross-attention.}
  \label{pic3}
  \vspace{-1.5em}
\end{figure*}

\subsection{Prompt Filtering Module}
\label{sec:Prompt filtering module}
\begin{figure}[h]
  \vspace{-0.5em}
  \centering
  \includegraphics[width=0.95\linewidth]{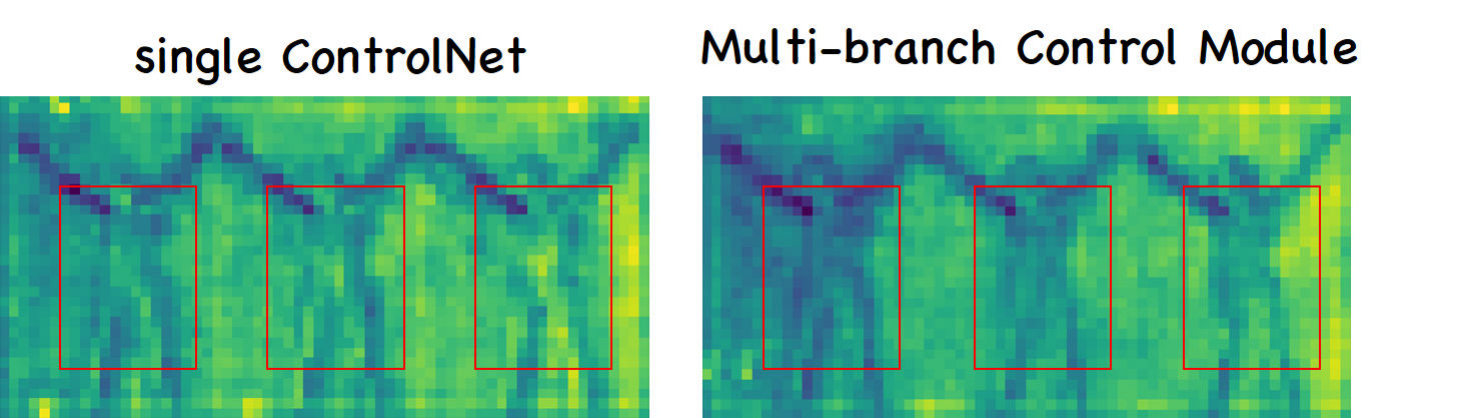}
  \caption{The visualization of feature maps based on different ControlNet types.}
  \label{pic4}
  \vspace{-0.1em}
\end{figure}

Prompt filtering module is designed to differentiate the description of different characters in prompt and project them separately. We use identifiers of the form: numbers enclosed in angle brackets, e.g. \textless1\textgreater,\textless2\textgreater, etc to identify the prompts. The exact display is on the left in Fig.~\ref{pic4}, we put \textless1\textgreater and \textless2\textgreater after "Trump" and "Biden" respectively, then a large language model split the prompt into  "Trump\textless1\textgreater, in a park" and "Biden\textless2\textgreater, in a park". According to the identifiers, the split prompts are assigned to different regions. Then we encode the tokenized prompts into text embeddings, and project them into $k$ and $v$ separately.

\subsection{Spatial Aligned Cross-attention}
\label{sec:spatial aligned Cross-attention}
Spatial aligned cross-attention is designed to generate each of the characters in prompt separately. Existing methods project hidden states as $q$, and implement attention calculation with $k$, $v$ in cross-attention, the formulation is as follows:
\begin{equation}
\begin{split}
Q=W_{Q}^{(i)}\cdot h_{i},K=W_{K}^{(i)}\cdot e_{i},V=W_{V}^{(i)}\cdot e_{i},& \\
h_{i+1}=\operatorname{Attention}\left(Q,K,V\right),&
\end{split}
\end{equation}

where $h_i$ and $h_{i+1}$ represent hidden states; $e_i$ represents text embeddings. In this process, prompt implements control of the generation. Here we find the crux of the problem: there is only one hidden states and one text embeddings being calculated in cross attention, which is a natural one-to-one process. There are no additional parameters to realize multiple control processes.

Considering this, we extend the structure of cross attention by copying the hidden states and entering them as a batch. The copied hidden states are projected as $q$ and implement attention calculation with $k$, $v$ from split prompts. In this way, we perform the one-to-one process several times in spatial aligned cross-attention in parallel. Descriptions of diverse characters in text prompts separately guide one cross attention process after that we obtain several hidden states corresponding to the split prompts.

Apart from this, we want to align the generated characters with the regions corresponding to the pose sequences. So we utilize the masks extracted from pose sequences to normalize the multiple hidden states. The complete process of spatial aligned cross-attention can be represented by the following formulation:
\begin{equation}
\begin{split}
h\, = \,\texttt{Attn}\left(Q_{0},K_{0},V_{0}\right) \cdot M_{0} \, + \, \texttt{Attn}\left(Q_{1},K_{1},V_{1}\right) \cdot M_{1} \, + \, \\ \texttt{Attn}\left(Q_{2},K_{2},V_{2}\right) \cdot M_{2} \, + \, \cdots \, + \, \texttt{Attn}\left(Q_{N},K_{N},V_{N}\right) \cdot M_{N},
\end{split}
\end{equation}

where $h$ represents the result of spatial aligned cross-attention module; $h_{i}$ represents the hidden states of diverse characters in text prompts corresponding to spatial section $i$; $m_i$ represents the masks extracted from the pose sequences in spatial section $i$.

\subsection{Multi-branch Control Module}
\label{sec:Multi-branch Control Module}
To gain a better alignment between split prompts and pose sequences in the same region, we design the multi-branch control module. Intuitively, split prompts and the corresponding pose sequences are processed in an individual ControlNet\cite{controlnet}, which avoids information contamination in the generation of each character. 

For example, in Fig.~\ref{pic2}, the input prompt is split according to the identifiers. The split embedding of "Trump\textless1\textgreater, in a park" and the pose sequences in the corresponding region are injected in the same ControlNet\cite{controlnet}, the other split embedding of "Biden\textless2\textgreater, in a park" and the pose sequences in the corresponding region are injected in another one. After these procedures, two characters "Trump" and "Biden" are generated separately in our module.

And then we utilize masks to bind the multiple branches. In details, we apply the manipulated masks to process the results of the multiple ControlNet\cite{controlnet}. This procedure is carried out in two groups and the formulations are demonstrated as below:
\begin{equation}
\begin{split}
S_{d}\, = \,S_{d_{0}} \cdot M_{d_{0}}\, + \,S_{d_{1}} \cdot M_{d_{1}}\, + \, \cdots \, + \,\,S_{d_{n}}\cdot M_{d_{N}},\\
S_{m}\, = \,S_{m_{0}} \cdot M_{m_{0}}\, + \,S_{m_{1}} \cdot M_{m_{1}}\, + \, \cdots \, + \,\,S_{m_{n}}\cdot M_{m_{N}},
\end{split}
\end{equation}

where $S_{d}$ and $S_{m}$ represent the results of multi-branch control module, $S_{di}$ and $S_{mi}$ separately denote the down block samples and the middle block samples of ControlNet\cite{controlnet}, and $S_{d}$ and $S_{di}$ are the lists containing 12 tensors categorized into four groups based on dimension. To be noticed, the first ControlNet\cite{controlnet} in this module is not processed by the masks, otherwise the quality of generation will experience a decline, which is evidenced by ablation experiments in Fig.~\ref{pic6}.

Toward the end, in Fig.~\ref{pic5}, we present the feature maps in cross attention of single ControlNet\cite{controlnet} and multi-branch control module. It is clear to observe that the feature map of our method is more cohesive and obvious, which further proves the effectiveness and necessity of this module.

%%实验部分删减了很多描述，是否合适

\begin{table*}[htbp]
\vspace{-3.0em}
\caption{Quantitative experiments with related works.}
\begin{center}
\begin{tabular*}{\linewidth}{@{\extracolsep{\fill}}ccccccccccccccc@{}}
    \toprule
    Method&Follow Your Pose&ControlVideo&Text2video-zero&FOMM&Masactrl+adapter&Tune-a-Video&Ours\\
    \midrule
    CS&31.13&29.67&29.02&28.19&31.77&29.32&\pmb{32.79}\\
    FC(\%)&96.32&96.07&96.21&79.03&88.27&94.37&\pmb{96.74}\\
    PA(\%)&34.17&32.35&32.19&9.7&33.91&26.54&\pmb{35.32}\\
    US(\%)&3.1&1.6&3.1&0&4.7&4.7&\pmb{82.8}\\
  \bottomrule
\end{tabular*}
\end{center}
\label{table1}
\vspace{-2.5em}
\end{table*}

\section{EXPERIMENTS}
\subsection{Implementation and Setting Details}
In our experiments, we employ Stable Diffusion V1.5 as the base model and we use DWPose\cite{dwpose} for the extraction of pose sequences. We further enhance base model with personalized Text-to-Image(T2I) checkpoints, including Realistic Vision, Toon You, Anything, Raemumix and HumanRealistic which allows for a variety of styles. Our experiments are conducted on one NVIDIA Tesla A800 GPU. 

\begin{figure}[h]
\vspace{-0.5em}
  \centering
  \includegraphics[width=0.97\linewidth]{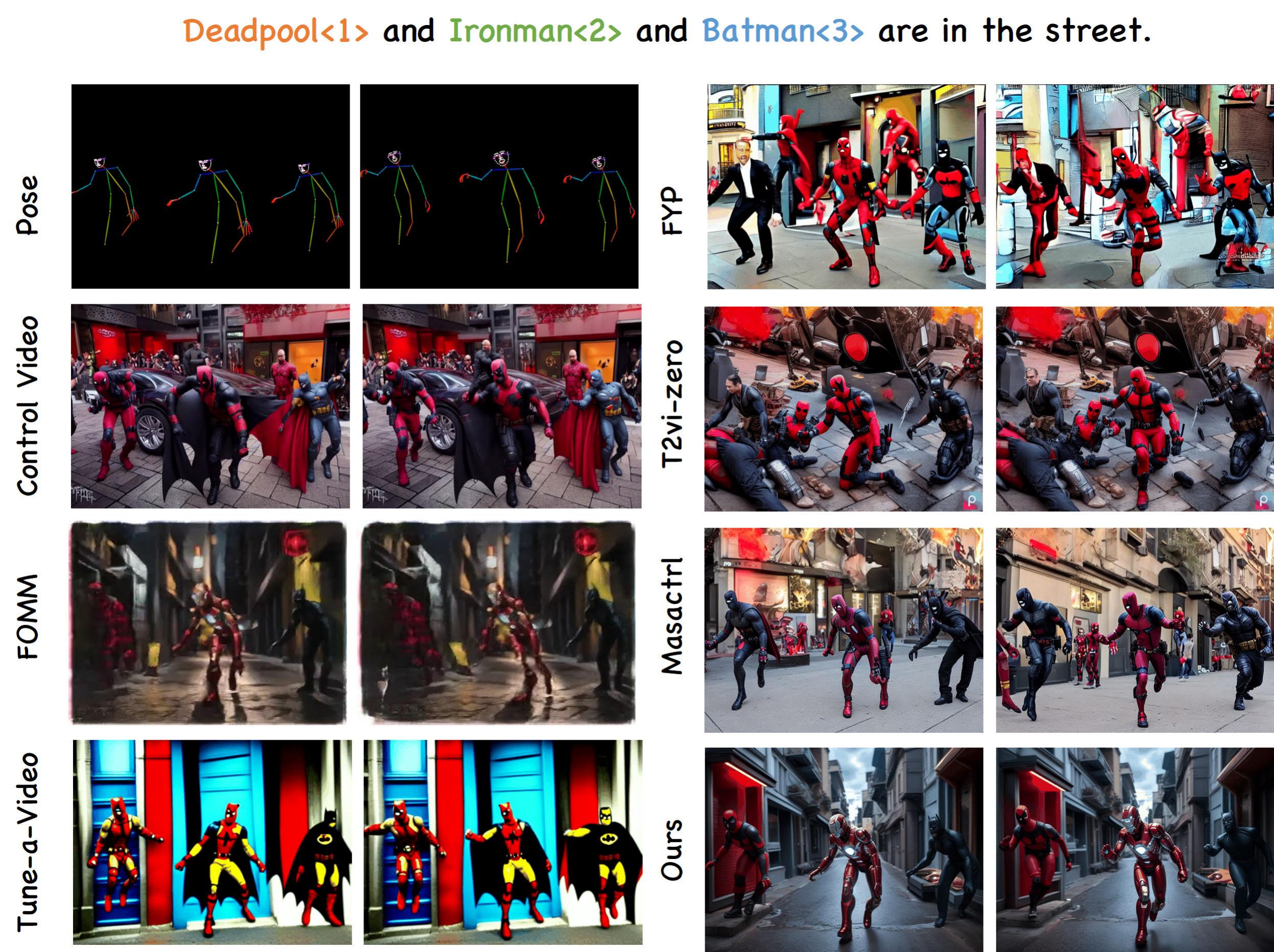}
  \caption{The illustration of generating videos of different approaches.}
  \label{pic5}
\vspace{-0.5em}
\end{figure}

\subsection{The Visualization of Generation Results.}
Our approach demonstrates various applications in generating videos with precise control. It facilitates the specification of each character during the generation process, yielding intriguing experimental outcomes. Furthermore, our approach allows for the manipulation of characters and backgrounds simply by adjusting the text prompt. Additionally, it supports various styles of video generation by loading different checkpoints downloaded from CivitAI\cite{civitai}. Some results are demonstrated in Fig.~\ref{pic1}.

\subsection{Comparison with Baselines}

%把实验合并一下，把表格扩展成长的

\textbf{Quantitative experiments.} We set up four metrics to compare the quality of the generation results, including CLIP score, frame consistency, pose accuracy, user study. Specifically, we calculate the CLIP score\cite{clipscore} to assess alignment between video and text and calculate CLIP cosine similarity between consecutive frames to assess frame consistency. By treating the input pose sequences as ground truth videos and average the precision results, we conduct an extensive evaluation of pose accuracy. Lastly, We conduct the user study with 20 participants. The results of experiments are shown in Table.~\ref{table1}.

\begin{figure}[h]
\vspace{-0.5em}
  \centering
  \includegraphics[width=0.97\linewidth]{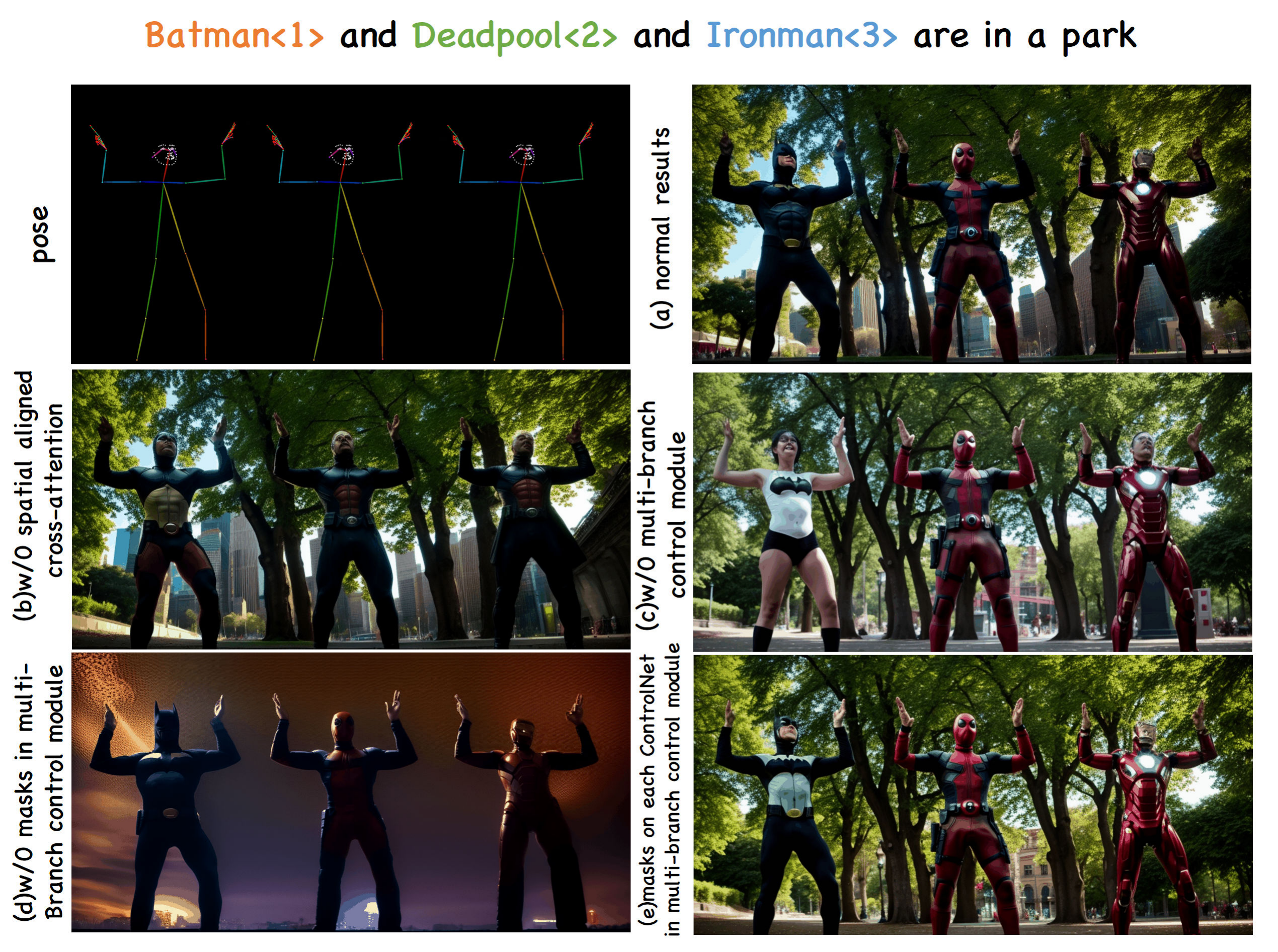}
  \caption{Ablation study, consists of parts (a) through parts (e). Part (a) demonstrates the generation results of complete model. Parts (b) and (c) respectively investigate the cases where the model is deprived of spatial aligned cross-attention and multi-branch control module. Parts (d) and (e) investigate the different fusion mechanisms within the multi-branch module.}
  \label{pic6}
\vspace{-0.5em}
\end{figure}

\textbf{Qualitative experiments.} In Fig.~\ref{pic5}, we compare our videos with those generated by other methods~\cite{follow,controlvideo,text2video,first,masactrl,t2i,tune}. Our approach showcased superior authenticity, consistency, and artistry, leading to better results.

\subsection{Consistency Experiments}
To prove the consistency of each character and background, we develop consistency experiments. In this paper, the experiment is conducted on videos that include two characters. Specifically, we design a mask extraction scheme and extract mask sequences of each character in the video. With these we can calculate the CLIP score of each character and background between consecutive frames, which can determine the level of consistency for each character and background. And we use these masks to calculate CLIP score between single character and the corresponding text prompts. The results of experiments are shown in Table.~\ref{table2}.

\begin{table}[h]
  \label{tab:freq}
\caption{The results of the consistency experiments.}
  \begin{tabular}{c c c c}
    \toprule
    Consistency&Character 1(\%)&Character 2(\%)&Background(\%)\\
    \midrule
    FC &\pmb{97.13} &\pmb{97.04} &\pmb{96.98} \\
    CS &\pmb{33.16} &\pmb{33.01} &\pmb{33.39} \\
  \bottomrule
\end{tabular}
\label{table2}
\vspace{-0.5em}
\end{table}

\subsection{Ablation Study}
To validate the effectiveness of our spatial aligned cross-attention and multi-branch control module, we conducted ablative experiments by removing spatial aligned cross-attention and merge multiple ControlNet\cite{controlnet} into one. To prove the necessity of the fusion mechanism in multi-branch control module, we remove and add masks at the results of each ControlNet\cite{controlnet}. The experimental results are shown in Fig.~\ref{pic6}, indicating that without the modules and methods we designed, the model’s multi-character generation performance would significantly deteriorate. 

\section{CONCLUSION}
In this paper, we use a tuning-free method to tackle the problem of multi-character generation via pose guidance. To achieve this, we firstly design mask enhancement flow to to obtain masks and locate the spatial section. Then, the prompt filtering module is designed to assign prompts of each character to its corresponding spatial section. The spatial aligned cross-attention aligns the prompts with the hidden states and generate characters separately. Lastly, we design multi-branch control module, which aligns the split prompts with each pose sequences. Leveraging our unique designs, we’re now capable of producing unique and imaginative multi-characters videos that are both temporally consistent and artistically compelling.

\section*{Acknowledgment}
This work was supported in part  by the Key Project of Technology Innovation and Application Development of Chongqing under Grant cstc2020jscx-gksbX0018, and in part by National Key R\&D Program of China under Grant 2018YFB2101200.


\begin{thebibliography}{00}
\bibitem{civitai} Civitai, ‘Civitai’, 2024. [Online]. Available: http://www.civitai.com/.

\bibitem{everybody} C. Chan, S. Ginosar, T. Zhou, and A. A. Efros, ‘Everybody dance now’, in Proceedings of the IEEE/CVF international conference on computer vision, 2019, pp. 5933–5942.

\bibitem{deep}
Y. Ren, G. Li, S. Liu, and T. H. Li, ‘Deep spatial transformation for pose-guided person image generation and animation’, IEEE Transactions on Image Processing, vol. 29, pp. 8622–8635, 2020.

\bibitem{bidirectionally}
W.-Y. Yu, L.-M. Po, R. C. C. Cheung, Y. Zhao, Y. Xue, and K. Li, ‘Bidirectionally deformable motion modulation for video-based human pose transfer’, in Proceedings of the IEEE/CVF International Conference on Computer Vision, 2023, pp. 7502–7512.

\bibitem{exploring}
P. Zhang, L. Yang, J.-H. Lai, and X. Xie, ‘Exploring dual-task correlation for pose guided person image generation’, in Proceedings of the IEEE/CVF Conference on Computer Vision and Pattern Recognition, 2022, pp. 7713–7722.

\bibitem{gan}
I. Goodfellow et al., ‘Generative adversarial networks’, Communications of the ACM, vol. 63, no. 11, pp. 139–144, 2020.

\bibitem{glide}
A. Nichol et al., ‘Glide: Towards photorealistic image generation and editing with text-guided diffusion models’, arXiv preprint arXiv:2112. 10741, 2021.

\bibitem{hierarchical}
A. Ramesh, P. Dhariwal, A. Nichol, C. Chu, and M. Chen, ‘Hierarchical text-conditional image generation with clip latents’, arXiv preprint arXiv:2204. 06125, vol. 1, no. 2, p. 3, 2022.

\bibitem{high}
R. Rombach, A. Blattmann, D. Lorenz, P. Esser, and B. Ommer, ‘High-resolution image synthesis with latent diffusion models’, in Proceedings of the IEEE/CVF conference on computer vision and pattern recognition, 2022, pp. 10684–10695.

\bibitem{photorealistic}
C. Saharia et al., ‘Photorealistic text-to-image diffusion models with deep language understanding’, Advances in neural information processing systems, vol. 35, pp. 36479–36494, 2022.

\bibitem{follow}
Y. Ma et al., ‘Follow your pose: Pose-guided text-to-video generation using pose-free videos’, in Proceedings of the AAAI Conference on Artificial Intelligence, 2024, vol. 38, pp. 4117–4125.

\bibitem{dreampose}
J. Karras, A. Holynski, T.-C. Wang, and I. Kemelmacher-Shlizerman, ‘Dreampose: Fashion video synthesis with stable diffusion’, in Proceedings of the IEEE/CVF International Conference on Computer Vision, 2023, pp. 22680–22690.

\bibitem{advancing}
F. Shen, H. Ye, J. Zhang, C. Wang, X. Han, and W. Yang, ‘Advancing pose-guided image synthesis with progressive conditional diffusion models’, arXiv preprint arXiv:2310. 06313, 2023.

\bibitem{humansd}
X. Ju, A. Zeng, C. Zhao, J. Wang, L. Zhang, and Q. Xu, ‘HumanSD: A native skeleton-guided diffusion model for human image generation’, in Proceedings of the IEEE/CVF International Conference on Computer Vision, 2023, pp. 15988–15998.

\bibitem{person}
A. K. Bhunia et al., ‘Person image synthesis via denoising diffusion model’, in Proceedings of the IEEE/CVF Conference on Computer Vision and Pattern Recognition, 2023, pp. 5968–5976.

%\bibitem{constitutional}
%Y. Bai et al., ‘Constitutional ai: Harmlessness from ai feedback’, %arXiv preprint arXiv:2212. 08073, 2022.

\bibitem{llama}
H. Touvron et al., ‘Llama 2: Open foundation and fine-tuned chat models’, arXiv preprint arXiv:2307. 09288, 2023.

%\bibitem{mt5}
%L. Xue et al., ‘mT5: A massively multilingual pre-trained text-to-text transformer’, arXiv preprint arXiv:2010. 11934, 2020.

%\bibitem{language}
%T. Brown et al., ‘Language models are few-shot learners’, Advances in neural information processing systems, vol. 33, pp. 1877–1901, 2020.

\bibitem{controlnet}
L. Zhang, A. Rao, and M. Agrawala, ‘Adding conditional control to text-to-image diffusion models’, in Proceedings of the IEEE/CVF International Conference on Computer Vision, 2023, pp. 3836–3847.

%\bibitem{dreambooth}
%N. Ruiz, Y. Li, V. Jampani, Y. Pritch, M. Rubinstein, and K. Aberman, ‘Dreambooth: Fine tuning text-to-image diffusion models for subject-driven generation’, in Proceedings of the IEEE/CVF Conference on Computer Vision and Pattern Recognition, 2023, pp. 22500–22510.

\bibitem{animatediff}
Y. Guo et al., ‘Animatediff: Animate your personalized text-to-image diffusion models without specific tuning’, arXiv preprint arXiv:2307. 04725, 2023.

%\bibitem{sketch}
%A. Voynov, K. Aberman, and D. Cohen-Or, ‘Sketch-guided text-to-image diffusion models’, in ACM SIGGRAPH 2023 Conference Proceedings, 2023, pp. 1–11.

%\bibitem{vdm}
%J. Ho, T. Salimans, A. Gritsenko, W. Chan, M. Norouzi, and D. J. Fleet, ‘Video diffusion models’, Advances in Neural Information Processing Systems, vol. 35, pp. 8633–8646, 2022.

%\bibitem{vldm}
%A. Blattmann et al., ‘Align your latents: High-resolution video synthesis with latent diffusion models’, in Proceedings of the IEEE/CVF Conference on Computer Vision and Pattern Recognition, 2023, pp. 22563–22575.

\bibitem{tune}
J. Z. Wu et al., ‘Tune-a-video: One-shot tuning of image diffusion models for text-to-video generation’, in Proceedings of the IEEE/CVF International Conference on Computer Vision, 2023, pp. 7623–7633.

%\bibitem{imagen}
%J. Ho et al., ‘Imagen video: High definition video generation with diffusion models’, arXiv preprint arXiv:2210. 02303, 2022.

\bibitem{text2video}
L. Khachatryan et al., ‘Text2video-zero: Text-to-image diffusion models are zero-shot video generators’, in Proceedings of the IEEE/CVF International Conference on Computer Vision, 2023, pp. 15954–15964.

%\bibitem{make}
%U. Singer et al., ‘Make-a-video: Text-to-video generation without text-video data’, arXiv preprint arXiv:2209. 14792, 2022.

\bibitem{controlvideo}
M. Zhao, R. Wang, F. Bao, C. Li, and J. Zhu, ‘Controlvideo: Adding conditional control for one shot text-to-video editing’, arXiv preprint arXiv:2305. 17098, 2023.

%\bibitem{disco}
%T. Wang et al., ‘Disco: Disentangled control for referring human dance generation in real world’, arXiv e-prints, p. arXiv-2307, 2023.

%\bibitem{magicanimate}
%Z. Xu et al., ‘Magicanimate: Temporally consistent human image animation using diffusion model’, arXiv preprint arXiv:2311. 16498, 2023.

%\bibitem{animate}
%L. Hu, X. Gao, P. Zhang, K. Sun, B. Zhang, and L. Bo, ‘Animate anyone: Consistent and controllable image-to-video synthesis for character animation’, arXiv preprint arXiv:2311. 17117, 2023.

\bibitem{t2i}
C. Mou et al., ‘T2i-adapter: Learning adapters to dig out more controllable ability for text-to-image diffusion models’, in Proceedings of the AAAI Conference on Artificial Intelligence, 2024, vol. 38, pp. 4296–4304.

%\bibitem{uni}
%S. Zhao et al., ‘Uni-controlnet: All-in-one control to text-to-image diffusion models’, Advances in Neural Information Processing Systems, vol. 36, 2024.

%\bibitem{unicontrol}
%C. Qin et al., ‘Unicontrol: A unified diffusion model for controllable visual generation in the wild’, arXiv preprint arXiv:2305. 11147, 2023.

%\bibitem{control-a-video}
%W. Chen et al., ‘Control-a-video: Controllable text-to-video generation with diffusion models’, arXiv preprint arXiv:2305. 13840, 2023.

%\bibitem{multidiffusion}
%O. Bar-Tal, L. Yariv, Y. Lipman, and T. Dekel, ‘Multidiffusion: Fusing diffusion paths for controlled image generation’, 2023.

%\bibitem{blip}
%D. Li, J. Li, and S. Hoi, ‘Blip-diffusion: Pre-trained subject representation for controllable text-to-image generation and editing’, Advances in Neural Information Processing Systems, vol. 36, 2024.

%\bibitem{rerender}
%S. Yang, Y. Zhou, Z. Liu, and C. C. Loy, ‘Rerender a video: Zero-shot text-guided video-to-video translation’, in SIGGRAPH Asia 2023 Conference Papers, 2023, pp. 1–11.

%\bibitem{magicvideo}
%D. Zhou, W. Wang, H. Yan, W. Lv, Y. Zhu, and J. Feng, ‘Magicvideo: Efficient video generation with latent diffusion models’, arXiv preprint arXiv:2211. 11018, 2022.

\bibitem{masactrl}
M. Cao, X. Wang, Z. Qi, Y. Shan, X. Qie, and Y. Zheng, ‘Masactrl: Tuning-free mutual self-attention control for consistent image synthesis and editing’, in Proceedings of the IEEE/CVF International Conference on Computer Vision, 2023, pp. 22560–22570.

\bibitem{first}
A. Siarohin, S. Lathuilière, S. Tulyakov, E. Ricci, and N. Sebe, ‘First order motion model for image animation’, Advances in neural information processing systems, vol. 32, 2019.

\bibitem{clipscore}
J. Hessel, A. Holtzman, M. Forbes, R. L. Bras, and Y. Choi, ‘Clipscore: A reference-free evaluation metric for image captioning’, arXiv preprint arXiv:2104. 08718, 2021.

\bibitem{gpt}
J. Achiam et al., ‘Gpt-4 technical report’, arXiv preprint arXiv:2303. 08774, 2023.

\bibitem{dwpose}
Z. Yang, A. Zeng, C. Yuan, and Y. Li, ‘Effective whole-body pose estimation with two-stages distillation’, in Proceedings of the IEEE/CVF International Conference on Computer Vision, 2023, pp. 4210–4220.

\bibitem{click}
Y. Ma et al., ‘Follow-Your-Click: Open-domain Regional Image Animation via Short Prompts’, arXiv preprint arXiv:2403. 08268, 2024.

\bibitem{stick}
Y. Ma et al., ‘MagicStick: Controllable Video Editing via Control Handle Transformations’, arXiv preprint arXiv:2312. 03047, 2023.

\bibitem{emoji}
Y. Ma et al., ‘Follow-Your-Emoji: Fine-Controllable and Expressive Freestyle Portrait Animation’, arXiv preprint arXiv:2406. 01900, 2024.

\bibitem{canvas}
Q. Chen et al., ‘Follow-Your-Canvas: Higher-Resolution Video Outpainting with Extensive Content Generation’, arXiv preprint arXiv:2409. 01055, 2024.

\bibitem{cove}
J. Wang, Y. Ma, J. Guo, Y. Xiao, G. Huang, and X. Li, ‘COVE: Unleashing the Diffusion Feature Correspondence for Consistent Video Editing’, arXiv preprint arXiv:2406. 08850, 2024.

\bibitem{swap}
C. Zhu et al., ‘InstantSwap: Fast Customized Concept Swapping across Sharp Shape Differences’, arXiv preprint arXiv:2412. 01197, 2024.

\bibitem{minds}
Y. Liu, Y. Ma, G. Zhu, H. Jing, and N. Zheng, ‘See Through Their Minds: Learning Transferable Neural Representation from Cross-Subject fMRI’, arXiv [cs.CV]. 2024.

\bibitem{dit}
K. Feng et al., ‘Dit4edit: Diffusion transformer for image editing’, arXiv preprint arXiv:2411. 03286, 2024.




%\bibitem{conditionvideo}
%B. Peng, X. Chen, Y. Wang, C. Lu, and Y. Qiao, ‘ConditionVideo: Training-Free Condition-Guided Video Generation’, in Proceedings of the AAAI Conference on Artificial Intelligence, 2024, vol. 38, pp. 4459–4467.

\end{thebibliography}
\end{document}